\documentclass[lettersize,journal]{IEEEtran}

\usepackage{amsmath,amssymb,amsfonts}
\usepackage{algorithm}
\usepackage{algorithmic}
\usepackage{array}
\usepackage{multirow}
\usepackage{graphicx}
\usepackage{textcomp}
\usepackage{stfloats}
\usepackage{url}
\usepackage{cite}
\usepackage{pifont}
\usepackage{booktabs}
\usepackage[table]{xcolor}
\usepackage[hidelinks]{hyperref}
\usepackage[caption=false,font=footnotesize]{subfig}

\begin{document}

\title{Consensus-Constrained Dual-Federated Optimization with Client-Side Dynamic Weighting for Distributed Android Device Threat Defense}

\author{Mohammad Mashreghi, Mohammad Hossein Badiei, and Md. Jalil Piran (Senior Member, IEEE)

\thanks{Mohammad Mashreghi and Mohammad H. Badiei are with the Department of Electrical and
Computer Engineering, University of Tehran, Tehran 1417614411, Iran, (email:m.mashreghi@ut.ac.ir; mho.badiei@gmail.com).}
\thanks{M. J. Piran is with the Department of Computer Science and Engineering,
Sejong University, Seoul 05006, South Korea, e-mail: piran@sejong.ac.kr}}

% The paper headers
% \markboth{Journal of \LaTeX\ Class Files,~Vol.~14, No.~8, August~2021}%
% {Shell \MakeLowercase{\textit{et al.}}: A Sample Article Using IEEEtran.cls for IEEE Journals}

% \IEEEpubid{0000--0000/00\$00.00~\copyright~2021 IEEE}
% Remember, if you use this you must call \IEEEpubidadjcol in the second
% column for its text to clear the IEEEpubid mark.

\maketitle

\begin{abstract}
Smartphones, smart-home hubs, and wearables are increasingly exposed to
sophisticated Android malware, while centralized threat-detection pipelines
raise privacy and scalability concerns. Federated learning (FL) enables
collaborative model training without collecting raw client data, but partial
participation and variation in local model quality can destabilize global
optimization. In this article, we propose DW-FedADMM, a federated optimization
framework that combines primal--dual coordination with performance-dependent
client-side weighting. The dynamic coefficient modulates each participating
client's local state before the augmented-Lagrangian update, while the dual
variables regulate disagreement between local and global model states. We
evaluate DW-FedADMM on three Android malware benchmarks, namely Drebin,
Malgenome, and TUANDROMD, with 5, 10, and 15 clients holding IID data
shards, partial client participation, and communication budgets of 5 and 10 rounds. Each
configuration is repeated with ten random seeds. DW-FedADMM obtains the
highest seed-averaged accuracy on Drebin ($97.64\%\pm0.20\%$) and the best
mean accuracy in 9 of the 18 dataset--client--round configurations,
including 7 of the 9 configurations with a 10-round budget. Pooled over all
configurations, it reaches $95.42\%$ accuracy, $94.62\%$ F1-score,
$96.06\%$ precision, and $99.45\%$ AUC, compared with $95.32\%$, $94.51\%$,
$95.56\%$, and $99.41\%$ for FedADMM, while also giving the lowest mean
false-positive rate and the smallest client update norm among the six
compared methods. Both primal--dual methods are well ahead of the averaging-
and proximal-based baselines on Malgenome and TUANDROMD, where the latter
often remain at the majority-class solution when clients are fragmented.
The gain of dynamic weighting over FedADMM is most pronounced on Drebin and
with the longer communication budget, where it also reduces the variation
across seeds.
\end{abstract}


\section{Introduction}

\IEEEPARstart{T}{he} proliferation of Android devices has revolutionized mobile computing, offering users unprecedented access to applications and services. However, this ubiquity has also made Android a prime target for malicious actors. The open-source nature of the Android platform, coupled with its fragmented ecosystem, has led to a surge in malware attacks, ranging from spyware and ransomware to sophisticated Advanced Persistent Threats (APTs)~\cite{almobaideen2025comprehensive}. Traditional malware-detection pipelines, built around centralized data collection, raise privacy concerns and fail to scale across the heterogeneous Android device landscape~\cite{cciplak2025fedetect, patel2025fl, ciaramella2026explainable}. 
Kaspersky's 2024 Mobile-Malware Evolution report logged 33.3 million blocked attacks involving mobile malware, adware, or unwanted software in a single year, illustrating the scale of the threat facing the Android ecosystem~\cite{Kaspersky2024}. Recent studies further highlight the associated privacy, security, and scalability challenges of on-device intelligence~\cite{pei2024review,rezaei2025survey}.

\begin{figure}[t]
\centering
{\includegraphics[width=0.5\textwidth]{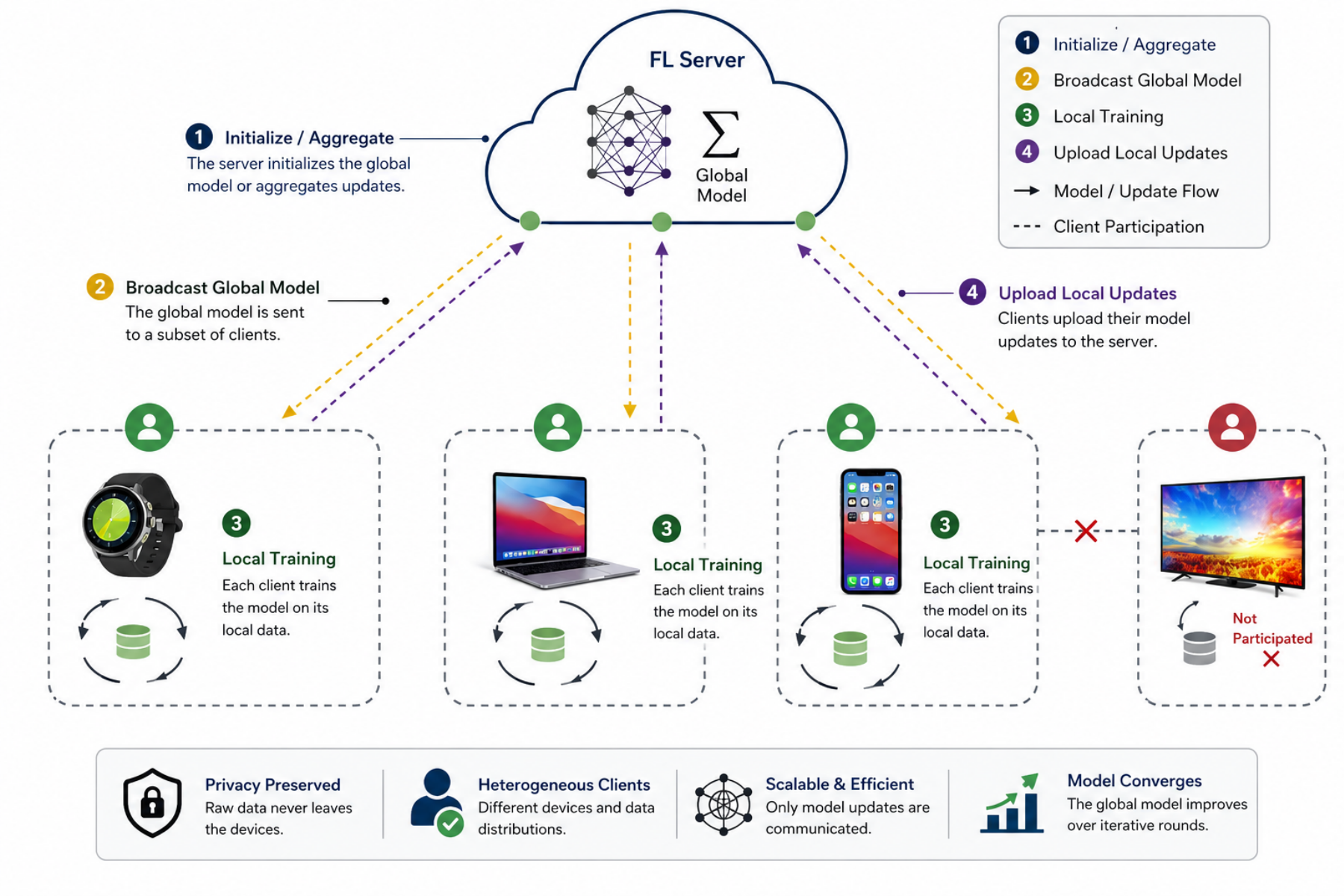}}
\caption{Overview of the federated learning framework and consumer-device ecosystem.}
\label{fig:FL}
\end{figure}

To address these challenges, federated learning (FL) has emerged as a promising paradigm for collaborative model training without centralizing raw client data \cite{mahmud2026survey}. As depicted in Fig.~\ref{fig:FL}, the server initializes or updates a global model and broadcasts it to a subset of available clients; each participating client then refines the model using its local data and returns an update for global coordination. This process is repeated until convergence or stabilization. FL-based security mechanisms have consequently been investigated across a broad range of consumer devices: blockchain-supported aggregation has been considered for smart-home hubs and routers~\cite{das2025blockchain,xia2024teva}, cross-device knowledge transfer has been used to address heterogeneity in wearables and in-car systems~\cite{chen2024trustworthy}, and privacy-preserving FL has been explored for smart-TV, IP-camera, and other IoT threat-detection scenarios~\cite{rabieinejad2024two,buyuktanir2025federated}.

Despite its privacy and scalability advantages, FL introduces optimization challenges that become particularly important in distributed malware detection. Frequent model synchronization incurs substantial communication overhead in bandwidth- and resource-constrained devices~\cite{jia2025comprehensive}. Differences in computational capability and intermittent connectivity can further produce delayed or asynchronous updates, while stochastic client availability means that only a subset of devices may participate in each communication round~\cite{kumari2024communication,garg2025client,mahmud2026survey}. Together, these effects can disrupt client--server coordination, slow convergence, and make the optimization process sensitive to the quality and history of individual client updates~\cite{thakur2024hardware}.

Several directions have been explored to improve the reliability of federated malware detection. Adversarial-resilient strategies based on GANs, robust optimization, and encrypted updates aim to protect training against malicious or unreliable participants~\cite{taheri2024unveiling,huang2024federated,eslamnejad2025federated,feng2025survey,li2025threats,jimenez2026security}. Byzantine-resilient aggregation mechanisms, including Krum and median-based approaches, provide additional robustness but can introduce scalability and computational limitations~\cite{zhu2026bpflh,khraisat2024survey,li2019rsa,xue2025enhancing}. At the model level, DNN- and CNN-based malware classifiers can capture complex patterns in permissions, API behavior, and network activity~\cite{augello2025m2fd,patel2025fl,xi2025gnndroid}; however, their computational demands can limit training and inference on devices with constrained CPU and memory resources~\cite{anand2025malite}. These limitations motivate optimization mechanisms that can maintain effective coordination without relying solely on more complex models or aggregation defenses.

Primal--dual optimization provides one such direction. Methods based on the Alternating Direction Method of Multipliers (ADMM) coordinate local and global model states through explicit consensus constraints and dual variables~\cite{boyd2011distributed}. This structure can limit persistent client--global disagreement, although maintaining dual states introduces additional memory and optimization steps and requires appropriate parameter selection~\cite{zhu2025fedapm}. FedADMM extends this formulation to federated optimization with partial client participation, allowing primal--dual coordination even when only a subset of clients is active in each round~\cite{wang2022fedadmm}. In parallel, performance-based weighting strategies such as Dynamic Weighted Federated Averaging (DW-FedAvg) use client performance to adapt the influence of local information during federated training~\cite{chaudhuri2023dynamic, wajahat2026dynamic}. Such strategies, however, are typically developed around averaging-based optimization and do not exploit the persistent local and dual states available in a primal--dual framework.

More recently, partial federated primal--dual optimization was applied to Android malware detection, showing that FedADMM coordination can provide effective classification under partial client participation~\cite{mashreghi2025partial}. Nevertheless, the client update in this formulation is initialized from the current global state without using previously observed client performance to determine how much information from the client's retained state should be preserved. This leaves a gap between performance-adaptive federated learning and primal--dual coordination: client-side predictive information is available during training, but is not used to adapt the starting point of the subsequent local primal--dual update. A mechanism that incorporates such information locally, while preserving the consensus objective and server-side aggregation of FedADMM, could provide greater continuity across intermittent client participation without altering the underlying coordination structure.

To address this gap, 
we propose \emph{Dual-Federated Optimization with Dynamic Weighting (DW-FedADMM)}\footnote{The source code is publicly available at \url{https://github.com/ArgAI-Lab/Dynamic-Weighted-Federated-Dual-Optimization}.}, which integrates performance-dependent client-side state modulation into consensus-constrained primal--dual federated optimization. Before each local augmented-Lagrangian update, a participating client interpolates between the current global model and its previously retained corrected state using a coefficient determined by its stored predictive performance relative to the best previously observed client performance. The resulting state is then used to initialize the standard primal update, followed by the dual correction and server aggregation. The dynamic coefficient therefore operates locally before optimization. Accordingly, the consensus formulation, augmented-Lagrangian client optimization, dual update, corrected-state transmission, and server aggregation follow the FedADMM-based primal--dual structure used in~\cite{wang2022fedadmm,mashreghi2025partial}, whereas the performance-dependent initialization rule constitutes the principal algorithmic extension.
The proposed method is evaluated against five federated baselines on Drebin, Malgenome, and TUANDROMD with 5, 10, and 15 clients, partial participation, and communication budgets of 5 and 10 rounds, with each configuration repeated over ten random seeds. DW-FedADMM achieves the highest overall mean accuracy (95.42\%) and the best accuracy in 9 of the 18 configurations, including 7 of 9 after 10 rounds. The primal--dual methods show a clear advantage over the averaging- and proximal-based baselines on Malgenome and TUANDROMD, while the additional benefit of dynamic weighting is most evident on Drebin and with the longer communication budget. Overall, the results suggest that performance-dependent state modulation primarily acts as a stabilizing complement to primal--dual coordination rather than providing uniform gains across all settings. The evaluation focuses on IID client partitions, which isolates the effect of client fragmentation and partial participation from that of statistical heterogeneity; extending the analysis to non-IID data is a natural next step.

% The source code for our experiments is publicly available on
% \href{https://github.com/ArgAI-Lab/Dynamic-Weighted-Federated-Dual-Optimization}{GitHub}.

The rest of the paper is organized as follows. Section~\ref{sec:methodology} details the proposed framework. Section~\ref{Simulation Results and Discussion} presents the experimental setup, results, and analysis. Section~\ref{sec:conclusion} concludes the paper, and Section~\ref{sec:future_work} outlines directions for future research.

\section{Methodology} \label{sec:methodology}

\subsection{Background} \label{Background}

To motivate the formulation of our proposed framework, we revisit the foundational components underlying federated mobile threat defense in Android ecosystems, emphasizing both system-level characteristics and optimization-theoretic principles.

\subsubsection{Android Application Structure} Android applications are distributed as APK packages that encapsulate critical execution logic and metadata. Each APK typically contains: (i) \texttt{AndroidManifest.xml}, specifying component declarations, runtime permissions, and app configuration; (ii) \texttt{classes.dex}, holding Dalvik Executable bytecode; and (iii) embedded resources (layouts, multimedia, and UI descriptors). These structured representations provide static and semantic features suitable for local inference. Within a federated setting, this modularity facilitates client-side feature extraction without sharing raw binaries, ensuring compliance with privacy constraints; this can be widely used across a wide range of Android devices, including smartphones, smartwatches, smart TVs, and home routers.

\subsubsection{Federated Optimization Framework} Consider \( n \) mobile clients indexed by \( i \in \{1, \dots, n\} \), each holding a local dataset \( \mathcal{D}_i \) with \( |\mathcal{D}_i| \) samples. The objective is to collaboratively optimize a global model parameterized by \( \bar{x} \in \mathbb{R}^d \) via:

\begin{equation}
\min_{\bar{x} \in \mathbb{R}^d} \; \sum_{i=1}^n \frac{|\mathcal{D}_i|}{|\mathcal{D}|} \, \mathbb{E}_{(u,y) \sim \mathcal{D}_i} \left[ \ell\big(h_{\bar{x}}(u), y\big) \right],
\end{equation}
where \( |\mathcal{D}|=\sum_{i=1}^{n}|\mathcal{D}_i| \), \( (u,y) \) is an input--label pair, \( \ell \) is a sample-wise loss function (e.g., cross-entropy), and \( h_{\bar{x}} \) denotes the model with parameters \( \bar{x} \). With equal-sized client shards, as in our experiments, the weights reduce to \( 1/n \). In classical FedAvg, each client solves:

\begin{equation}
x_i^k \leftarrow \operatorname{SGD}^{E}\left( \bar{x}^{k-1}; \mathcal{D}_i \right), \quad \text{followed by} \quad \bar{x}^k \leftarrow \frac{1}{n} \sum_{i=1}^{n} x_i^k,
\end{equation}
where \( \mathrm{SGD}_E \) denotes \( E \) local minibatch SGD steps in round \( k \). Despite its simplicity, FedAvg suffers in heterogeneous environments due to unconstrained client deviation.

To address this, FedProx augments the local objective with a proximal regularization:

\begin{equation}
x_i^k = \arg\min_{x \in \mathbb{R}^d} \left\{ \mathcal{L}_i(x) + \frac{\mu}{2} \| x - \bar{x}^{k-1} \|^2 \right\},
\end{equation}
where \( \mu > 0 \) controls proximity to the global reference point \( \bar{x}^{k-1} \). This penalization mitigates client drift, especially under heterogeneous settings, and enhances convergence stability.

% \subsubsection{Federated Mobile Threat Defense} 
% \textcolor{red}{
\subsubsection{Federated Mobile Threat Defense in Consumer Devices} 

Leveraging federated optimization for Android threat defense enables scalable threat modeling without centralizing sensitive behavioral data. Clients extract high-dimensional representations from static code features, dynamic system traces, permission graphs, and execution profiles. These vectors inform distributed training of a mobile threat classifier. However, federated deployment introduces unique complications: (i) communication bottlenecks due to iterative synchronization, (ii) statistical heterogeneity across local datasets, and (iii) temporal asynchrony from fluctuating client availability. These issues impair coordination and motivate dual-informed optimization together with performance-adaptive client mechanisms, as explored in this work.

% Ultimately, all baseline algorithms that employ dynamic weighting were adapted to incorporate client-side dynamic weighting, following the same strategy as our method.

\subsection{Proposed Method}
\label{sec:method}

\begin{figure}[t]
\centering
\includegraphics[width=0.5\textwidth]
{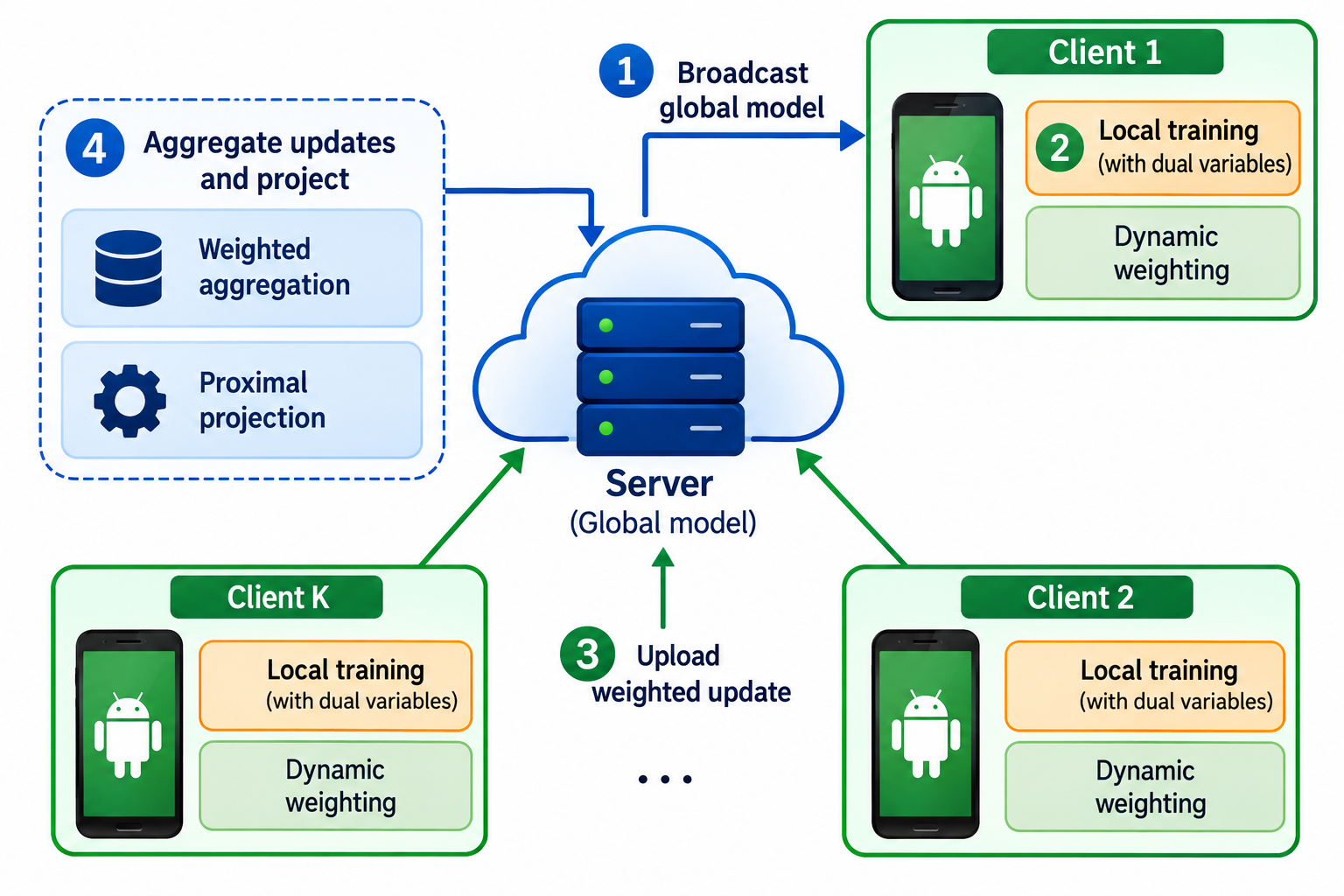}
\caption{Overview of the proposed DW-FedADMM framework for federated Android
malware detection.}
\label{fig:DW-FedDual}
\end{figure}

We propose DW-FedADMM, a primal--dual federated optimization framework that
integrates consensus regulation with performance-dependent client-side
modulation. As illustrated in Fig.~\ref{fig:DW-FedDual}, the framework
combines an augmented-Lagrangian formulation with dynamic weighting of the
local model state prior to each client update. The resulting procedure is
designed for federated threat classification under partial client
participation while retaining explicit coordination between local and global
model variables.

\subsubsection{Problem Decomposition}

Consider a federated threat-detection system consisting of \(n\) clients,
where client \(i\in[n]\) holds a private dataset \(\mathcal{D}_i\).
Each client defines a local empirical objective \(f_i(x)\), corresponding to
the classification loss evaluated on its locally available malware samples.
Because the training partition is divided into approximately equal-sized
client shards in the reported experiments, we adopt the equal-client
objective
\begin{equation}
\min_x\;
F(x)
=
\frac{1}{n}
\sum_{i=1}^{n} f_i(x).
\label{eq:federated_objective}
\end{equation}
A local variable \(x_i\) is associated with each client, and consensus with
the server model \(\bar{x}\) is enforced through
\begin{equation}
x_i=\bar{x},
\qquad \forall i\in[n],
\label{eq:consensus_constraint}
\end{equation}
where \(\bar{x}\) denotes the global model maintained by the server.

\subsubsection{Augmented-Lagrangian Formulation}

The corresponding consensus-constrained problem is written as
\begin{equation}
\min_{\{x_i\},\,\bar{x}}
\frac{1}{n}\sum_{i=1}^{n} f_i(x_i),
\qquad
\mathrm{s.t.}\quad
x_i=\bar{x},
\quad \forall i\in[n].
\label{eq:consensus_problem}
\end{equation}
Introducing a dual variable \(z_i\) for each client and a quadratic consensus
penalty yields the client-level augmented Lagrangian
\begin{equation}
\mathcal{L}_i(x_i,\bar{x},z_i)
=
f_i(x_i)
+
\left\langle z_i,x_i-\bar{x}\right\rangle
+
\frac{\gamma}{2}
\left\|x_i-\bar{x}\right\|_2^2,
\label{Augmented Lagrangian}
\end{equation}
where \(\gamma>0\) controls the strength of the consensus penalty. The
linear dual term accounts for accumulated disagreement between the local and
global states, whereas the quadratic term penalizes instantaneous deviation
from the current global model.

\subsubsection{Dynamic Weight Modulation}

DW-FedADMM introduces a performance-dependent coefficient that modulates the
client state before local optimization. At the beginning of communication
round \(k\), the server keeps the most recently reported client-side
accuracy \(\mathrm{Acc}_{j,\mathrm{prev}}\) of every client \(j\) in the set
\(\mathcal{O}_k\subseteq[n]\) of clients that have participated at least once,
with \(\mathcal{O}_0=\emptyset\). The reference score is the largest of these
stored values,
\begin{equation}
\mathrm{BestAcc}_{k}
=
\max_{j\in\mathcal{O}_k}
\mathrm{Acc}_{j,\mathrm{prev}}.
\label{eq:best_accuracy_reference}
\end{equation}
Each participating client \(i\) then computes
\begin{equation}
\omega_{i,k}
=
\begin{cases}
\dfrac{\mathrm{Acc}_{i,\mathrm{prev}}}
{\max\left(\mathrm{BestAcc}_{k},\epsilon\right)},
& i\in\mathcal{O}_k,\\[2mm]
1, & i\notin\mathcal{O}_k,
\end{cases}
\label{Dynamic Weight}
\end{equation}
where \(\epsilon>0\) is a small constant that keeps the ratio defined.
Unobserved clients therefore do not enter the reference maximum, and no
artificial score is assigned to them. A client that participates for the
first time uses \(\omega_{i,k}=1\), so its local training starts from the
current global model; in the first round this holds for every selected
client. Because \(\mathrm{BestAcc}_{k}\) is the largest score among the
observed clients, \(0\leq\omega_{i,k}\leq1\).

The resulting coefficient is used to combine the current global model with
the client's previously corrected state,
\begin{equation}
x_i^{k}
\leftarrow
\omega_{i,k}\bar{x}^{k}
+
\left(1-\omega_{i,k}\right)\hat{x}_i^{k}.
\label{Coefficient Dynamic Weight}
\end{equation}
Hence, the dynamic coefficient controls the relative contribution of current
global information and the previously retained client state before the
subsequent primal update. The coefficient is therefore a client-side
modulation factor rather than a direct server-side aggregation weight.

The direction of this interpolation follows from what a low score indicates
in the primal--dual setting. A client with a low
\(\mathrm{Acc}_{i,\mathrm{prev}}\) relative to \(\mathrm{BestAcc}_{k}\) is
one whose local data are not yet well represented by the consensus model. Its
corrected state \(\hat{x}_i^{k}\) contains both the local progress made on
these data and, through \(z_i^{k}/\eta\), the accumulated disagreement with
the global model. Resetting such a client fully to \(\bar{x}^{k}\) would
discard this information and force it to repeat the same local correction in
every round. Retaining a larger share of \(\hat{x}_i^{k}\) instead gives a
warm start that keeps the client's trajectory continuous across rounds, while
the dual term in \eqref{Augmented Lagrangian} still pulls the local model
toward consensus. Clients with high relative scores are already well served
by the global model, so starting them close to \(\bar{x}^{k}\) costs little
and keeps them aligned. The rule is therefore a heuristic relaxation of the
local starting point, not a property derived from the objective, and its
value is assessed empirically by comparing DW-FedADMM with FedADMM, which
corresponds to \(\omega_{i,k}=1\) for all clients.

\subsubsection{Primal--Dual Client Update}

Starting from the modulated local state in
\eqref{Coefficient Dynamic Weight}, each selected client runs a finite number
of SGD steps on its local augmented Lagrangian,
\begin{equation}
\begin{aligned}
x_{i,0}^{k} &= \omega_{i,k}\bar{x}^{k}+\left(1-\omega_{i,k}\right)\hat{x}_i^{k},\\
x_{i,e+1}^{k} &= x_{i,e}^{k}-\alpha\,\widehat{\nabla}\mathcal{L}_i\!\left(x_{i,e}^{k},\bar{x}^{k},z_i^{k}\right),
\quad e=0,\dots,E-1,\\
x_i^{k+1} &= x_{i,E}^{k},
\end{aligned}
\label{eq:primal_update}
\end{equation}
where \(\widehat{\nabla}\) denotes a minibatch stochastic gradient, \(\alpha\)
is the local learning rate, and \(E\) is the total number of minibatch steps
taken over the local epochs of one round.
The corresponding dual variable is then updated according to
\begin{equation}
z_i^{k+1}
=
z_i^{k}
+
\eta
\left(
x_i^{k+1}-\bar{x}^{k}
\right).
\label{eq:dual_update}
\end{equation}
The updated primal and dual variables are subsequently combined to construct
the corrected local state
\begin{equation}
\hat{x}_i^{k+1}
=
x_i^{k+1}
+
\frac{1}{\eta}z_i^{k+1},
\label{eq:corrected_state}
\end{equation}
where \(\eta>0\) is the dual step size, which is also used to scale the
dual correction, while \(\gamma\) only weights the quadratic penalty in
\eqref{Augmented Lagrangian}. Each participating
client transmits \(\hat{x}_i^{k+1}\), together with its local validation accuracy \(\mathrm{Acc}_{i,k}\), to the server. Clients that are
not selected in round \(k\) keep their dual variable, corrected state, and
stored score unchanged until they participate again; the local primal
variable is not stored between rounds, since it is rebuilt from
\eqref{Coefficient Dynamic Weight} at the start of each participation.

\subsubsection{Server Aggregation}

After receiving updates from the participating subset
\(S_k\subseteq[n]\), the server replaces the stored accuracy of each
participating client by \(\mathrm{Acc}_{i,k}\) and sets
\(\mathcal{O}_{k+1}=\mathcal{O}_k\cup S_k\); the stored values of inactive
clients are retained. The reference score for the next round is then
\begin{equation}
\mathrm{BestAcc}_{k+1}
=
\max_{j\in\mathcal{O}_{k+1}}
\mathrm{Acc}_{j,\mathrm{prev}}.
\label{eq:best_accuracy_update}
\end{equation}
The corrected client states received in round \(k\) are aggregated according
to
\begin{equation}
\tilde{x}^{k+1}
=
\frac{1}{|S_k|}
\sum_{i\in S_k}
\hat{x}_i^{k+1}.
\label{eq:server_aggregation}
\end{equation}
For completeness, the server update can be written as
\begin{equation}
\bar{x}^{k+1}
=
\operatorname{prox}_{(\gamma/n)g}
\left(
\tilde{x}^{k+1}
\right),
\label{eq:global_proximal_update}
\end{equation}
where \(g\) denotes an optional server-side regularizer. In the reported
experiments no additional server-side regularizer is applied
(\(g\equiv0\)), and therefore the proximal mapping reduces to the identity,
i.e., \(\bar{x}^{k+1}=\tilde{x}^{k+1}\).

\begin{algorithm}[t]
\caption{DW-FedADMM}
\label{alg:fedadmm}
\begin{algorithmic}[1]

\STATE \textbf{Initialize:}
$x^0$, $\gamma>0$, $\eta>0$, and number of rounds $K$
\STATE Set $\bar{x}^{0}\leftarrow x^0$
\STATE Set $x_i^0\leftarrow x^0$, $z_i^0\leftarrow0$,
and $\hat{x}_i^0\leftarrow x^0$ for all $i\in[n]$
\STATE Set $\mathcal{O}_0\leftarrow\emptyset$ (no stored client scores)

\FOR{$k=0,1,\ldots,K-1$}

    \STATE Randomly sample $S_k\subseteq[n]$ with $|S_k|=\lceil 0.8\,n\rceil$
    \STATE $\mathrm{BestAcc}_{k}\leftarrow
    \max_{j\in\mathcal{O}_k}\mathrm{Acc}_{j,\mathrm{prev}}$ (if $\mathcal{O}_k\neq\emptyset$)

    \STATE \textbf{$\vartriangleright$ Client-side update}

    \FOR{each $i\in S_k$}

        \STATE Receive $\bar{x}^{k}$ and $\mathrm{BestAcc}_{k}$

        \STATE
        $\omega_{i,k}
        \leftarrow
        \mathrm{Acc}_{i,\mathrm{prev}}/\max(\mathrm{BestAcc}_{k},\epsilon)$
        if $i\in\mathcal{O}_k$, otherwise $\omega_{i,k}\leftarrow1$

        \STATE
        $x_i^{k}
        \leftarrow
        \omega_{i,k}\bar{x}^{k}
        +
        (1-\omega_{i,k})\hat{x}_i^{k}$

        \STATE
        Run $E$ SGD steps on $\mathcal{L}_i(\cdot,\bar{x}^{k},z_i^{k})$
        from $x_i^{k}$ to obtain $x_i^{k+1}$ (Eq.~\eqref{eq:primal_update})

        \STATE
        $z_i^{k+1}
        \leftarrow
        z_i^{k}
        +
        \eta
        \left(
        x_i^{k+1}-\bar{x}^{k}
        \right)$

        \STATE
        $\hat{x}_i^{k+1}
        \leftarrow
        x_i^{k+1}
        +
        \dfrac{1}{\eta}z_i^{k+1}$

        \STATE Compute $\mathrm{Acc}_{i,k}$ on the client's local validation subset
        \STATE Send $\hat{x}_i^{k+1}$ and $\mathrm{Acc}_{i,k}$ to the server

    \ENDFOR

    \STATE \textbf{for each} $i\notin S_k$: keep $z_i$, $\hat{x}_i$, and
    $\mathrm{Acc}_{i,\mathrm{prev}}$ unchanged

    \STATE \textbf{$\vartriangleright$ Server-side update}

    \FOR{each $i\in S_k$}
        \STATE $\mathrm{Acc}_{i,\mathrm{prev}}\leftarrow\mathrm{Acc}_{i,k}$
    \ENDFOR

    \STATE $\mathcal{O}_{k+1}\leftarrow\mathcal{O}_k\cup S_k$ and
    $\mathrm{BestAcc}_{k+1}
    \leftarrow
    \max_{j\in\mathcal{O}_{k+1}}
    \mathrm{Acc}_{j,\mathrm{prev}}$

    \STATE
    $\tilde{x}^{k+1}
    \leftarrow
    \dfrac{1}{|S_k|}
    \displaystyle
    \sum_{i\in S_k}
    \hat{x}_i^{k+1}$

    \STATE
    $\bar{x}^{k+1}
    \leftarrow
    \operatorname{prox}_{(\gamma/n)g}
    \left(
    \tilde{x}^{k+1}
    \right)$

\ENDFOR

\end{algorithmic}
\end{algorithm}

\subsubsection{Optimization Protocol}

Algorithm~\ref{alg:fedadmm} summarizes the round-wise interaction between
the clients and the server. At each communication round, the server selects
a subset of clients and provides them with the current global model and the
reference performance value. Each participating client first applies the
performance-dependent modulation in \eqref{Coefficient Dynamic Weight},
runs a fixed number of SGD steps on its local augmented Lagrangian, updates the
associated dual variable, and returns the corrected local state together with
its current client-side accuracy.

The server updates the stored performance records, computes
\(\mathrm{BestAcc}_{k+1}\), aggregates the corrected states of the
participating clients, and applies the global update in
\eqref{eq:global_proximal_update}. The resulting global model is then used in
the next communication round. In this manner, DW-FedADMM combines
performance-dependent local adaptation with primal--dual coordination under
partial client participation.

\subsubsection{Dynamically Weighted Baselines}
DW-FedAvg and DW-FedProx use the same coefficient \(\omega_{i,k}\) in
\eqref{Dynamic Weight}, but without dual variables. Each selected client
starts its local training from
\begin{equation}
x_{i,0}^{k}=\omega_{i,k}\bar{x}^{k}+\left(1-\omega_{i,k}\right)x_i^{\mathrm{prev}},
\label{eq:dw_baselines}
\end{equation}
where \(x_i^{\mathrm{prev}}\) is the local model the client obtained at its
last participation. DW-FedAvg then runs local SGD on \(f_i\), and DW-FedProx
on \(f_i(x)+\frac{\mu}{2}\|x-\bar{x}^{k}\|^2\) with \(\mu=0.01\). In both
cases the server takes the plain average of the returned local models. With
\(\omega_{i,k}=1\), the two methods reduce exactly to FedAvg and FedProx.

\section{Simulation Results and Discussion} \label{Simulation Results and Discussion}

To assess the proposed method, we conduct simulations across a range of client populations and communication budgets, and every configuration is repeated with ten random seeds. The analysis focuses on predictive performance, client-fragmentation behavior, false-positive rates, convergence over rounds, and optimization-update statistics relative to established FL baselines.

% \subsection{Dataset Description}
\label{sec:dataset_description}

To evaluate the proposed approach, We use three benchmark datasets for Android malware detection: Drebin, Malgenome, and TUANDROMD.

\vspace{2mm}\noindent\textbf{Drebin:} Following the benchmark configuration used in
prior federated Android-malware studies~\cite{chaudhuri2023dynamic}, the
Drebin-derived dataset contains 15,036 application samples, with 9,476 benign
and 5,560 malicious instances, represented by 215 features.

\vspace{2mm}\noindent\textbf{Malgenome:} This dataset, derived from the Android Malware
Genome Project~\cite{yerima2018droidfusion}, comprises 3,799 Android
application samples, including 2,539 benign and 1,260 malicious applications.
It contains 215 extracted features.

\vspace{2mm}\noindent\textbf{TUANDROMD:} The TUANDROMD benchmark described
in~\cite{borah2020malware} contains 4,465 Android application instances and
241 features for binary malware/goodware classification. The experiments use
the released preprocessed feature representation.

\subsection{Experimental Setup}
\label{sec:experiments}

This subsection summarizes the computational environment, base classifier,
federated partitioning protocol, and optimization settings used in the
experiments.

\subsubsection{Machine Configuration}
The experiments were run on a single GPU-equipped machine. All methods share
the same code base and data pipeline.

\subsubsection{Base Classifier and Optimizer}
All federated methods use the same four-layer feed-forward neural network for
binary classification. The three hidden layers contain 200, 100, and 50
neurons with ReLU activation, followed by a single sigmoid output neuron.
The architecture is intentionally kept fixed across all methods so that the
comparison primarily reflects differences in federated optimization rather
than classifier capacity. The model has 68,401 parameters for Drebin and
Malgenome (215 features) and 73,601 for TUANDROMD (241 features). Weights use
the default PyTorch initialization of linear layers, and training minimizes
the binary cross-entropy loss on the sigmoid output. Local parameters are
updated with plain SGD, without momentum, weight decay, or gradient clipping.
All methods are implemented in PyTorch.

\subsubsection{Training Parameters}
Each dataset is divided using an 80--20 train--test hold-out split. The
training partition is randomly shuffled and divided into equal-sized client
shards, so the client data are IID: every client sees approximately the same
class ratio and feature distribution. No Dirichlet, label-skew, or
quantity-skew partition is used. Increasing the number of clients therefore
reduces the number of samples available to each client, but it does not make
the local distributions more different from each other. Throughout the
results, we use \emph{client fragmentation} only in this sense, that is, more
clients with smaller local datasets, and not as a measure of statistical
heterogeneity. At every
communication round, 80\% of the available clients are selected uniformly
for participation.

The training configuration is:
\begin{itemize}
    \item \textit{Batch Size}: 32.
    \item \textit{Local Epochs}: 8 per participating client.
    \item \textit{Learning Rate}: \(10^{-2}\) for local SGD.
    \item \textit{Client Populations}: 5, 10, and 15 clients.
    \item \textit{Communication Budgets}: each run lasts 10 rounds, and the
    global model is evaluated after round 5 and after round 10. These two
    checkpoints give the 5- and 10-round budgets reported in the results.
    \item \textit{Client Selection}: 0.8 of the available clients per round.
    \item \textit{Primal--Dual Parameters}: the consensus/dual scaling
    parameters are fixed across all configurations, with
    \(\gamma=\eta=0.01\). FedProx and DW-FedProx use a proximal coefficient
    of \(0.01\).
    \item \textit{Random Seeds}: ten seeds shared by all methods.
\end{itemize}

For the adaptive coefficient, each client holds out a fixed 20\% of its
training shard as a local validation subset, drawn only from the training
partition, and trains on the remaining 80\%. After local training, the client
measures the classification accuracy of its updated local model, before the
dual correction, on this validation subset, using the standard decision
threshold of the sigmoid output. This accuracy is the score that the client
reports to the server. The score is stored by the server and reused until that client participates again; clients that
have not yet participated have no stored score, as described in
Section~\ref{sec:method}. The held-out global test partition is used only for
global-model evaluation and is not required by the dynamic-weight formula.

For the optimization analysis, the client update norm is measured as the
Euclidean distance \(\|x_i^{k+1}-\bar{x}^{k}\|_2\) between a client's
updated local parameters and the current global parameters, computed over
the trainable weights before the dual correction and averaged over the
participating clients of a round. A smaller norm alone does not imply
stability; we read it only together with the accuracy results. The false-positive rate is computed from the test confusion matrix as
\(\mathrm{FP}/(\mathrm{FP}+\mathrm{TN})\), with malware as the positive class.

\subsection{Discussion and Analysis}

\begin{table*}[t]
\centering
\caption{Global-model accuracy across datasets, client populations, and communication budgets (in \%). Values are mean $\pm$ standard deviation over ten random seeds, evaluated after 5 and 10 communication rounds. The best mean for each row and budget is shown in bold. All methods use the same ten seeds.}
\label{tab:Global-Models-Metrics-global-acc-16epoch-5-10-rounds}

\setlength{\tabcolsep}{3.8pt}
\renewcommand{\arraystretch}{1.5}
\scriptsize

\resizebox{\textwidth}{!}{%
\begin{tabular}{
ll
cccccc
cccccc
}
\toprule

\multirow{2}{*}{\textbf{Dataset}}
&
\multirow{2}{*}{\textbf{Clients}}
&
\multicolumn{6}{c}{\textbf{5 Communication Rounds}}
&
\multicolumn{6}{c}{\textbf{10 Communication Rounds}}
\\

\cmidrule(lr){3-8}
\cmidrule(lr){9-14}

&
&
\textbf{FedAvg}
&
\textbf{FedProx}
&
\textbf{FedADMM}
&
\textbf{DW-FedAvg}
&
\textbf{DW-FedProx}
&
\textbf{DW-FedADMM}
&
\textbf{FedAvg}
&
\textbf{FedProx}
&
\textbf{FedADMM}
&
\textbf{DW-FedAvg}
&
\textbf{DW-FedProx}
&
\textbf{DW-FedADMM}
\\

\midrule

\multirow{3}{*}{\textbf{Drebin}}
& 5
& \textbf{97.69 $\pm$ 0.18}
& 97.65 $\pm$ 0.17
& 97.29 $\pm$ 1.23
& \textbf{97.69 $\pm$ 0.16}
& 97.66 $\pm$ 0.19
& 97.63 $\pm$ 0.39
& 98.25 $\pm$ 0.27
& 98.22 $\pm$ 0.27
& 97.08 $\pm$ 2.44
& \textbf{98.29 $\pm$ 0.19}
& 98.25 $\pm$ 0.22
& 98.12 $\pm$ 0.56
\\

& 10
& 96.66 $\pm$ 0.26
& 96.66 $\pm$ 0.25
& 97.57 $\pm$ 0.38
& 96.64 $\pm$ 0.25
& 96.64 $\pm$ 0.24
& \textbf{97.65 $\pm$ 0.23}
& 97.70 $\pm$ 0.25
& 97.69 $\pm$ 0.26
& 97.78 $\pm$ 0.96
& 97.70 $\pm$ 0.24
& 97.68 $\pm$ 0.23
& \textbf{97.95 $\pm$ 0.45}
\\

& 15
& 95.39 $\pm$ 0.47
& 95.38 $\pm$ 0.48
& \textbf{97.06 $\pm$ 0.27}
& 95.37 $\pm$ 0.41
& 95.35 $\pm$ 0.41
& 97.04 $\pm$ 0.23
& 97.22 $\pm$ 0.22
& 97.22 $\pm$ 0.21
& 95.72 $\pm$ 2.18
& 97.18 $\pm$ 0.21
& 97.17 $\pm$ 0.21
& \textbf{97.42 $\pm$ 1.00}
\\

\midrule

\rowcolor{gray!7}

& 5
& 96.28 $\pm$ 1.45
& 96.22 $\pm$ 1.45
& \textbf{97.99 $\pm$ 0.42}
& 96.05 $\pm$ 1.49
& 95.95 $\pm$ 1.42
& 97.95 $\pm$ 0.42
& 98.07 $\pm$ 0.24
& 98.07 $\pm$ 0.24
& 98.70 $\pm$ 0.39
& 98.07 $\pm$ 0.28
& 98.07 $\pm$ 0.28
& \textbf{98.72 $\pm$ 0.43}
\\

\rowcolor{gray!7}

& 10
& 66.84 $\pm$ 0.00
& 66.84 $\pm$ 0.00
& \textbf{96.78 $\pm$ 1.00}
& 66.84 $\pm$ 0.00
& 66.84 $\pm$ 0.00
& 96.53 $\pm$ 1.53
& 96.36 $\pm$ 1.28
& 96.33 $\pm$ 1.31
& 98.41 $\pm$ 0.36
& 96.11 $\pm$ 1.42
& 96.11 $\pm$ 1.42
& \textbf{98.45 $\pm$ 0.30}
\\

\rowcolor{gray!7}
\multirow{-3}{*}{\textbf{Malgenome}}
& 15
& 66.84 $\pm$ 0.00
& 66.84 $\pm$ 0.00
& \textbf{86.43 $\pm$ 13.24}
& 66.84 $\pm$ 0.00
& 66.84 $\pm$ 0.00
& 85.70 $\pm$ 12.75
& 89.51 $\pm$ 8.25
& 89.11 $\pm$ 9.23
& \textbf{97.97 $\pm$ 0.47}
& 88.36 $\pm$ 9.68
& 88.33 $\pm$ 9.66
& 97.93 $\pm$ 0.52
\\

\midrule

\multirow{3}{*}{\textbf{TUANDROMD}}
& 5
& 87.49 $\pm$ 6.12
& 87.17 $\pm$ 6.05
& \textbf{97.41 $\pm$ 0.67}
& 87.31 $\pm$ 6.10
& 86.98 $\pm$ 6.03
& 97.39 $\pm$ 0.65
& 98.03 $\pm$ 0.33
& 98.02 $\pm$ 0.31
& 98.28 $\pm$ 0.25
& 98.03 $\pm$ 0.33
& 98.01 $\pm$ 0.35
& \textbf{98.29 $\pm$ 0.29}
\\

& 10
& 79.84 $\pm$ 0.00
& 79.84 $\pm$ 0.00
& \textbf{86.72 $\pm$ 6.19}
& 79.84 $\pm$ 0.00
& 79.84 $\pm$ 0.00
& 86.24 $\pm$ 6.29
& 88.28 $\pm$ 5.87
& 88.10 $\pm$ 5.92
& 97.79 $\pm$ 0.38
& 87.63 $\pm$ 5.97
& 87.54 $\pm$ 6.04
& \textbf{97.85 $\pm$ 0.33}
\\

& 15
& 79.84 $\pm$ 0.00
& 79.84 $\pm$ 0.00
& \textbf{79.87 $\pm$ 0.07}
& 79.84 $\pm$ 0.00
& 79.84 $\pm$ 0.00
& \textbf{79.87 $\pm$ 0.07}
& 79.85 $\pm$ 0.04
& 79.85 $\pm$ 0.04
& 96.88 $\pm$ 0.59
& 79.85 $\pm$ 0.04
& 79.85 $\pm$ 0.04
& \textbf{96.91 $\pm$ 0.57}
\\
\bottomrule
\end{tabular}%
}
\end{table*}

In this subsection, we analyze the predictive performance, scalability, false-alarm
behavior, and optimization characteristics of DW-FedADMM relative to FedAvg,
FedProx, FedADMM, DW-FedAvg, and DW-FedProx. The evaluation covers three
Android malware datasets (Drebin, Malgenome, and TUANDROMD), client
populations of 5, 10, and 15, and communication budgets of 5 and 10 rounds,
with 8 local epochs and partial client participation. All reported values are
taken over ten random seeds. The configuration-level accuracy is given in
Table~\ref{tab:Global-Models-Metrics-global-acc-16epoch-5-10-rounds}, and the
figures below summarize the main trends across datasets and operating
conditions.

\begin{figure}[!h]
    \centering
    \includegraphics[width=0.45\textwidth]
    {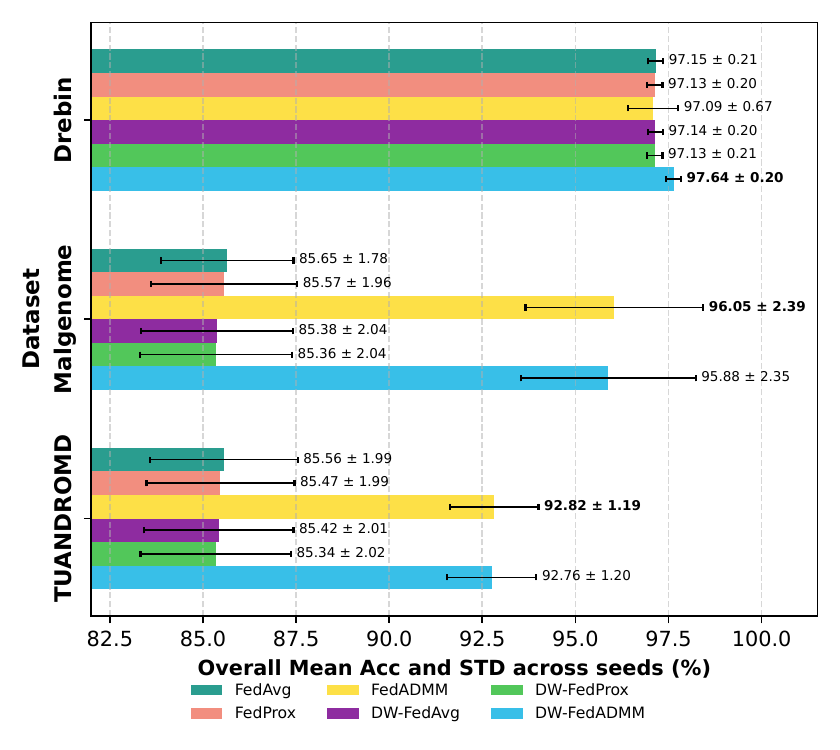}
    \caption{Overall mean global accuracy for each dataset. For each seed,
    accuracy is averaged over the three client populations and the two
    communication budgets; bars show the mean over ten seeds and error bars
    the standard deviation across seeds.}
    \label{fig:overall_mean_accuracy}
\end{figure}

\noindent\textbf{Dataset-level and aggregate accuracy.}
Figure~\ref{fig:overall_mean_accuracy} shows that the ranking of the methods
depends on the dataset. On Drebin, DW-FedADMM has the highest mean accuracy,
$97.64\%\pm0.20\%$, ahead of FedAvg ($97.15\%\pm0.21\%$) and FedADMM
($97.09\%\pm0.67\%$), and its spread across seeds is about one third of that
of FedADMM. On Malgenome and TUANDROMD, the two primal--dual methods are far
ahead of the others. FedADMM reaches $96.05\%$ and $92.82\%$ and DW-FedADMM
$95.88\%$ and $92.76\%$, while the four averaging- and proximal-based methods
stay between $85.3\%$ and $85.7\%$ on both datasets. Pooled over all 18
configurations, DW-FedADMM has the highest mean accuracy ($95.42\%$),
followed closely by FedADMM ($95.32\%$), whereas FedAvg, FedProx, DW-FedAvg,
and DW-FedProx lie between $89.27\%$ and $89.45\%$.

The configuration-level results in
Table~\ref{tab:Global-Models-Metrics-global-acc-16epoch-5-10-rounds} give a
more detailed picture. DW-FedADMM achieves the highest mean accuracy in 9 of the 18 configurations, comprising eight sole wins and one shared win with FedADMM on TUANDROMD.
The only settings won by an averaging-based method are Drebin with 5 clients, where FedAvg and DW-FedAvg share the best result after 5 rounds, and DW-FedAvg is marginally ahead after 10 rounds. The communication budget also affects the ranking: with 10 rounds, DW-FedADMM is best in 7 of the 9 configurations, whereas with 5 rounds FedADMM is best in 6 of the 9 and shares the best result in one more.
The gap
between the primal--dual and the averaging-based methods is largest when
clients are fragmented and the budget is short. With 10 or 15 clients and 5
rounds, FedAvg, FedProx, DW-FedAvg, and DW-FedProx predict only the majority
class on Malgenome ($66.84\%$) and TUANDROMD ($79.84\%$) in every seed, which
is why their standard deviation is zero. On TUANDROMD with 15 clients they
are still at $79.85\%$ after 10 rounds, while FedADMM and DW-FedADMM reach
$96.88\%$ and $96.91\%$.

\noindent\textbf{Comparison with FedADMM.}
Because FedADMM and DW-FedADMM are run on the same seeds and differ only in
the dynamic coefficient, their comparison isolates the effect of this
coefficient. DW-FedADMM is ahead of FedADMM in 10 of the 18 configurations.
Averaged over all configurations of a dataset, the gain is $+0.55$ points on
Drebin, while the two methods are nearly equal on Malgenome ($-0.17$ points)
and TUANDROMD ($-0.07$ points). The largest gains appear on Drebin after 10
rounds with 5 and 15 clients ($+1.03$ and $+1.70$ points). In these settings
FedADMM loses accuracy in some seeds between round 5 and round 10, whereas
DW-FedADMM does not. The dynamic coefficient therefore acts mainly as a
stabilizing component that strengthens primal--dual coordination where it is
most needed.

\begin{figure}[t]
    \centering
    \includegraphics[width=0.5\textwidth]
    {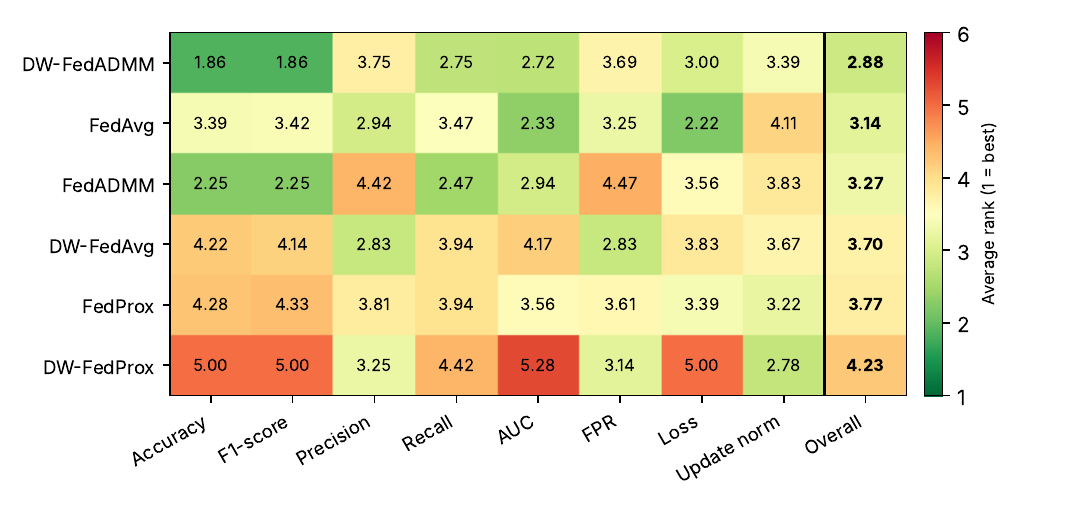}
    \caption{Average rank of the six methods over the 18 dataset--client--round
    configurations (1 = best). For each configuration, the methods are ranked
    on the seed-averaged value of each metric; lower is better for FPR, loss,
    and update norm. The last column is the mean rank over all eight metrics,
    and rows are sorted by it.}
    \label{fig:average_rank}
\end{figure}

\noindent\textbf{Overall ranking across metrics.}
To summarize the comparison in a single view, Fig.~\ref{fig:average_rank}
ranks the six methods in every configuration and averages the ranks over the
18 configurations. DW-FedADMM has the best average rank for accuracy and
F1-score (both $1.86$, against $2.25$ for FedADMM and $3.39$ or more for the
other methods), and it also has the best overall rank when all eight metrics
are combined ($2.88$), ahead of FedAvg ($3.14$) and FedADMM ($3.27$).
Compared with FedADMM, it ranks better on seven of the eight metrics,
including precision, FPR, loss, and update norm, and is slightly worse only
on recall.
FedAvg ranks better on AUC and loss; the averaging-based methods rank better on precision and FPR in many configurations because, on Drebin and Malgenome, their false alarms are very low; and the FedProx variants often rank better on update norm, although DW-FedADMM has the smallest mean update norm overall.
 The pooled mean FPR still favors DW-FedADMM, since it is
dominated by the large errors of the averaging-based methods on TUANDROMD.
The ranks describe how often and by how much a method is ahead; as shown
above, the clearest gain over FedADMM is observed on Drebin.

The communication budget matters for this picture. With 10 rounds,
DW-FedADMM is ahead of FedADMM in all three client populations ($98.38\%$
versus $98.02\%$, $98.08\%$ versus $97.99\%$, and $97.42\%$ versus $96.86\%$
for 5, 10, and 15 clients, averaged over the three datasets) and has the
smaller spread across seeds in each case. With 5 rounds the two methods are
within $0.25$ points of each other, and FedADMM is slightly ahead with 10 and
15 clients. The dynamic coefficient therefore seems to need a few rounds of
client feedback before it pays off.

\begin{figure*}[t]
    \centering
    \subfloat[Drebin]{\includegraphics[width=0.32\textwidth]{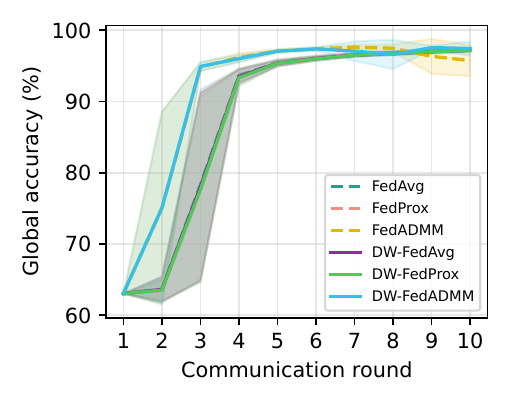}\label{fig:roundwise_15_drebin}}\hfill
    \subfloat[Malgenome]{\includegraphics[width=0.32\textwidth]{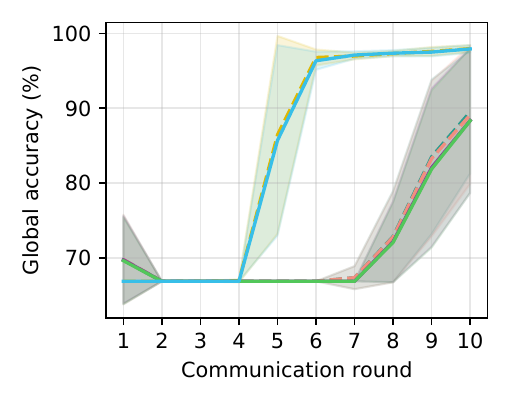}\label{fig:roundwise_15_malgenome}}\hfill
    \subfloat[TUANDROMD]{\includegraphics[width=0.32\textwidth]{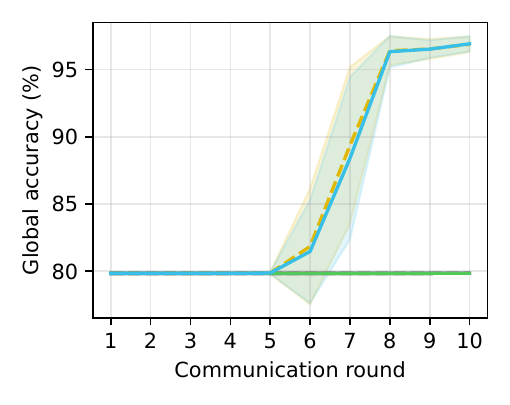}\label{fig:roundwise_15_tuandromd}}
    \caption{Round-wise global accuracy with 15 clients. Lines show the mean
    over ten seeds and shaded bands one standard deviation. Dashed lines
    denote the unweighted methods and solid lines their dynamically weighted
    counterparts.}
    \label{fig:roundwise_15}
\end{figure*}

\noindent\textbf{Convergence over rounds.}
Figure~\ref{fig:roundwise_15} gives a round-by-round view of the most
fragmented setting. On Malgenome and TUANDROMD, all methods start at the
majority-class solution. FedADMM and DW-FedADMM leave it after four to five
rounds on Malgenome and after six to seven rounds on TUANDROMD, and then
climb above $96\%$ within two or three more rounds. FedAvg and FedProx, with
or without dynamic weighting, reach only $88$--$90\%$ on Malgenome by round
10 and do not leave the majority class on TUANDROMD within the budget. On
Drebin, the ADMM-based methods pass $94\%$ by round 3, two rounds earlier
than the averaging-based methods. FedADMM then falls from $97.6\%$ to
$95.7\%$ over the last three rounds, while DW-FedADMM ends at $97.4\%$.
Dynamic weighting changes the FedAvg and FedProx curves very little.

\noindent\textbf{False-alarm behavior under fragmentation.}
False-positive rate is particularly important in malware detection because
it measures the fraction of benign applications that are incorrectly flagged
as malicious. Pooled over the three datasets, both budgets, and ten seeds,
DW-FedADMM has the lowest mean FPR of the six methods at $10.86\%$, followed
by FedADMM at $11.07\%$, while FedAvg, FedProx, DW-FedAvg, and DW-FedProx lie
between $24.1\%$ and $24.4\%$. FPR rises with the number of clients for every
method, but far more slowly for the primal--dual ones. With 5 clients,
DW-FedADMM reaches $2.03\%$ against $2.51\%$ for FedADMM and about $11.5\%$
for the other four methods. With 10 clients FedADMM is marginally lower
($12.18\%$ versus $12.52\%$), and with 15 clients DW-FedADMM is lower again
($18.02\%$ versus $18.52\%$), whereas the averaging- and proximal-based
methods climb to about $27\%$ and $34\%$. Most of these false alarms come
from TUANDROMD. Its test split contains only 180 benign applications, and a
model that has not yet left the majority (malware) class flags all of them.
On Drebin and Malgenome the mean FPR stays below $3\%$ for every method, and
on Drebin DW-FedADMM lowers it from $2.84\%$ to $1.83\%$ compared with
FedADMM.

\begin{figure}[t]
    \centering
    \includegraphics[width=0.4\textwidth]
    {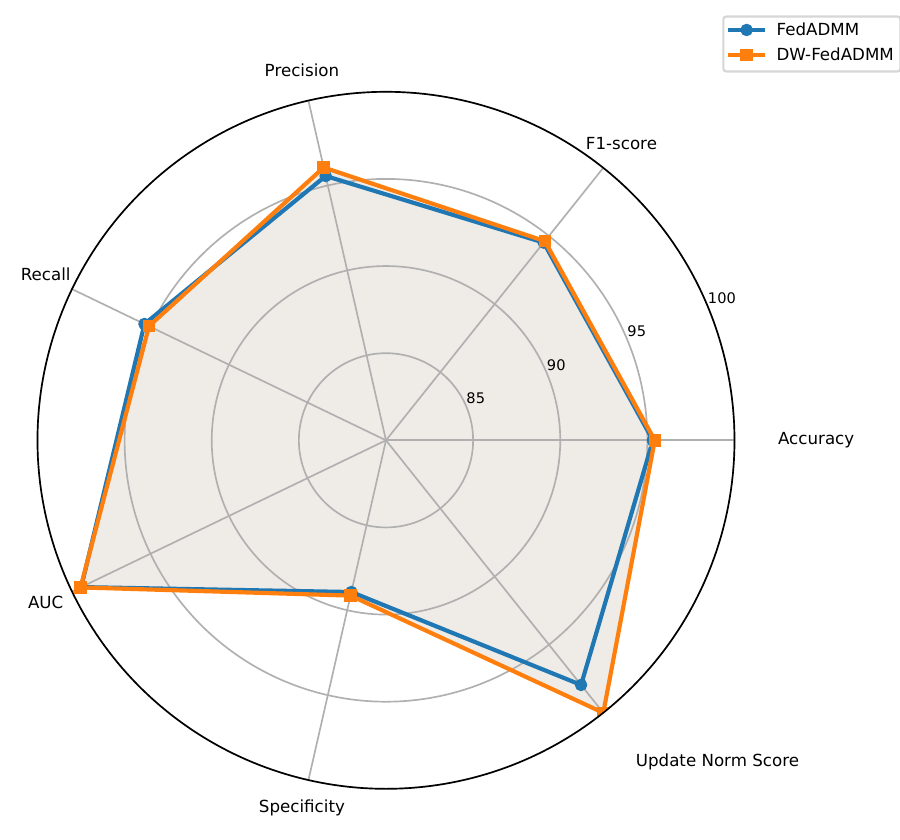}
    \caption{Integrated comparison of FedADMM and DW-FedADMM. Accuracy,
    F1-score, precision, recall, AUC, and specificity ($100-\mathrm{FPR}$)
    are overall means in percent. The update norm is shown as a
    lower-is-better score relative to the smallest mean update norm among
    the six methods, so that higher values are preferable. The radial axis
    starts at 80 to make the differences visible.}
    \label{fig:fedadmm_dwfedadmm_relative}
\end{figure}

\noindent\textbf{Integrated predictive and optimization comparison.}
Figure~\ref{fig:fedadmm_dwfedadmm_relative} compares the two primal--dual
methods on several criteria at once. Pooled over all 18 configurations and
ten seeds, DW-FedADMM is slightly ahead in accuracy ($95.42\%$ versus
$95.32\%$), F1-score ($94.62\%$ versus $94.51\%$), precision ($96.06\%$
versus $95.56\%$), AUC ($99.45\%$ versus $99.41\%$), and specificity
($89.14\%$ versus $88.93\%$), while FedADMM has a slightly higher recall
($95.38\%$ versus $95.09\%$). The two polygons largely overlap, consistent with the comparison above. The dynamic coefficient moves the operating
point a little toward higher precision and fewer false alarms, at the cost of
a small loss in recall, rather than changing the overall level of
performance.

DW-FedADMM also has the smallest mean update norm of all six methods
($0.2097$), slightly below FedADMM ($0.2140$) and clearly below the
averaging- and proximal-based methods ($0.228$--$0.231$). On the relative
scale, it therefore receives a score of $100$ against $98.0$ for FedADMM.
The difference between the two primal--dual methods is small, but it points
in the expected direction: the client-side coefficient slightly reduces how
far local models move away from the global model in each round.

\noindent\textbf{Loss interpretation and empirical variability.}
Binary cross-entropy loss should be interpreted separately from accuracy,
F1-score, AUC, and FPR. The mean test loss of DW-FedADMM ($0.245$) is about
$23\%$ lower than that of FedADMM ($0.318$) and close to the values of
FedAvg, FedProx, and their weighted variants ($0.239$--$0.244$), even though
those methods are much less accurate on Malgenome and TUANDROMD. This is not
contradictory because cross-entropy depends on the confidence and
calibration of predicted probabilities, whereas F1-score and FPR depend on
the selected decision threshold and AUC primarily evaluates ranking quality.
Accordingly, the results do not support a claim that DW-FedADMM minimizes
classification loss, although it clearly reduces the loss of FedADMM.

The multi-seed evaluation also shows where the results are fragile. The
largest standard deviations occur in short-budget, fragmented settings, such
as Malgenome with 15 clients after 5 rounds ($\pm12.75$ for DW-FedADMM and
$\pm13.24$ for FedADMM) and TUANDROMD with 10 clients after 5 rounds
($\pm6.29$ and $\pm6.19$), where some seeds leave the majority-class
solution earlier than others. A single run in these settings can therefore
be misleading, and our conclusions therefore rest on the seed-averaged values rather than on
individual runs.

\noindent\textbf{Overall implication.}
Taken together, the results support a more qualified conclusion than a
single-run evaluation would suggest. The primal--dual structure is the main
source of improvement. FedADMM and DW-FedADMM are much more accurate than the
averaging- and proximal-based methods on Malgenome and TUANDROMD, converge
faster in fragmented settings, and cut the mean false-positive rate by more
than half. Adding the dynamic coefficient on top of FedADMM gives the best
overall mean accuracy, F1-score, precision, and FPR among the six methods,
the smallest update norm, the clearest improvement on Drebin, and smaller
variation across seeds when 10 rounds are available. On Malgenome and
TUANDROMD, where FedADMM already performs strongly, the two methods are
nearly equal, and the benefit of the coefficient grows with the
communication budget. When the same coefficient is added to
FedAvg or FedProx, it has almost no effect. DW-FedADMM is therefore best
viewed as a robust choice for fragmented clients with a reasonable
communication budget, not as a method that dominates every metric and
dataset.
\section{Conclusion}
\label{sec:conclusion}

In this paper, we introduced DW-FedADMM, a federated optimization framework that
combines primal--dual coordination with performance-dependent client-side
modulation for distributed Android malware detection. The method was
evaluated on Drebin, Malgenome, and TUANDROMD under different client
populations and communication budgets, and every configuration was repeated
over ten random seeds. DW-FedADMM achieved the highest overall mean
accuracy, F1-score, and precision, the lowest mean false-positive rate, and
the smallest update norm among the six compared methods, and it was the best
method in 9 of the 18 configurations and in 7 of the 9 configurations with a
10-round budget. The multi-seed analysis also showed that most of the
advantage over averaging-based federated learning comes from primal--dual
coordination, while the dynamic coefficient adds a smaller, dataset-dependent
gain. This gain was most evident on Drebin, where it was also accompanied by lower
variation across seeds.
Overall, the results indicate that combining client-side performance
feedback with primal--dual coordination is a reliable choice under partial
client participation and small local datasets, provided that enough
communication rounds are available. Because all partitions in this study are
IID, these findings should not be read as evidence about performance under
non-IID client data, which remains to be tested.

\section{Future Work}
\label{sec:future_work}

The most immediate next step is to evaluate DW-FedADMM under non-IID client
data. This includes label-skew Dirichlet partitions with different
concentration parameters, quantity skew, clients that hold different malware
families, and temporal partitions where collection dates are available, with
client class ratios and a quantitative heterogeneity measure reported for
each setting. Since the present experiments use IID shards only, such an
evaluation is needed to test the heterogeneity-related motivation of the
method.

Beyond this, future work will extend DW-FedADMM toward a more adaptive
closed-loop federated controller. One direction is to replace the current scalar
accuracy-based coefficient with a stateful client-reliability measure that
combines predictive quality, score freshness, consensus residuals, and local
update dynamics. This reliability state could jointly adapt the client-side
modulation factor and the primal--dual parameters $\gamma$ and $\eta$,
allowing statistical reliability and consensus enforcement to evolve
together. A second direction is event-triggered communication, in which
client updates are transmitted only when their expected optimization benefit
exceeds a communication-dependent threshold. Finally, integrating change-point
detection with the reliability state could help distinguish transient client
degradation from genuine malware-distribution drift, enabling continual
adaptation to emerging threat patterns without discarding previously learned
client and dual states.

\bibliography{references}

\end{document}